\documentclass[10pt,twocolumn,letterpaper]{article}

\usepackage{cvpr}              

\definecolor{cvprblue}{rgb}{0.21,0.49,0.74}
\usepackage[pagebackref,breaklinks,colorlinks,allcolors=cvprblue]{hyperref}
\usepackage{multirow}
\usepackage[accsupp]{axessibility}

\def\paperID{39} 
\def\confName{CVPR}
\def\confYear{2025}

\title{CaddieSet: A Golf Swing Dataset with Human Joint Features and Ball Information}

\author{
Seunghyeon Jung$^{1}$ \quad
Seoyoung Hong$^{2}$ \quad
Jiwoo Jeong$^{1}$ \quad
Seungwon Jeong$^{1}$ \quad \\
Jaerim Choi$^{2}$ \quad
Hoki Kim$^{3,*}$ \quad
Woojin Lee$^{1,*}$ \quad
\\ \\
$^{1}$Dongguk University, Seoul, Republic of Korea \quad
$^{2}$Kimcaddie Inc, Seoul, Republic of Korea \\
$^{3}$Chung-Ang University, Seoul, Republic of Korea \\
$^{1}${\tt\small \{alzkdpf23, anpan8, youai058, wj926\}@dgu.ac.kr}, \\
$^{2}${\tt\small \{seoyoung, jrim\}@kimcaddie.com}, \quad
$^{3}${\tt\small hokikim@cau.ac.kr}
}

\begin{document}
\maketitle

\renewcommand{\thefootnote}{\fnsymbol{footnote}}
\footnotetext[1]{Corresponding authors.}

\begin{abstract}
Recent advances in deep learning have led to more studies to enhance golfers' shot precision. However, these existing studies have not quantitatively established the relationship between swing posture and ball trajectory, limiting their ability to provide golfers with the necessary insights for swing improvement. In this paper, we propose a new dataset called CaddieSet, which includes joint information and various ball information from a single shot. 
CaddieSet extracts joint information from a single swing video by segmenting it into eight swing phases using a computer vision-based approach. Furthermore, based on expert golf domain knowledge, we define 15 key metrics that influence a golf swing, enabling the interpretation of swing outcomes through swing-related features. Through experiments, we demonstrated the feasibility of CaddieSet for predicting ball trajectories using various benchmarks. In particular, we focus on interpretable models among several benchmarks and verify that swing feedback using our joint features is quantitatively consistent with established domain knowledge. This work is expected to offer new insight into golf swing analysis for both academia and the sports industry.
\end{abstract}    
\section{Introduction}
\label{1}

Golf is a sport that emphasizes precision and technique, with the ultimate goal for golfers being to hit the ball accurately and far \cite{hume2005role}. In the golf world, there is a saying that "the ball trajectory doesn't lie", which means that the flight path of the ball directly reflects the outcome of the swing. Therefore, continuous swing correction is essential for achieving accurate shots, not only for professional players but also for amateur golfers \cite{smith2012professional}.

\begin{figure*}[htb!]
    \centering
    \includegraphics[width=0.8\linewidth]{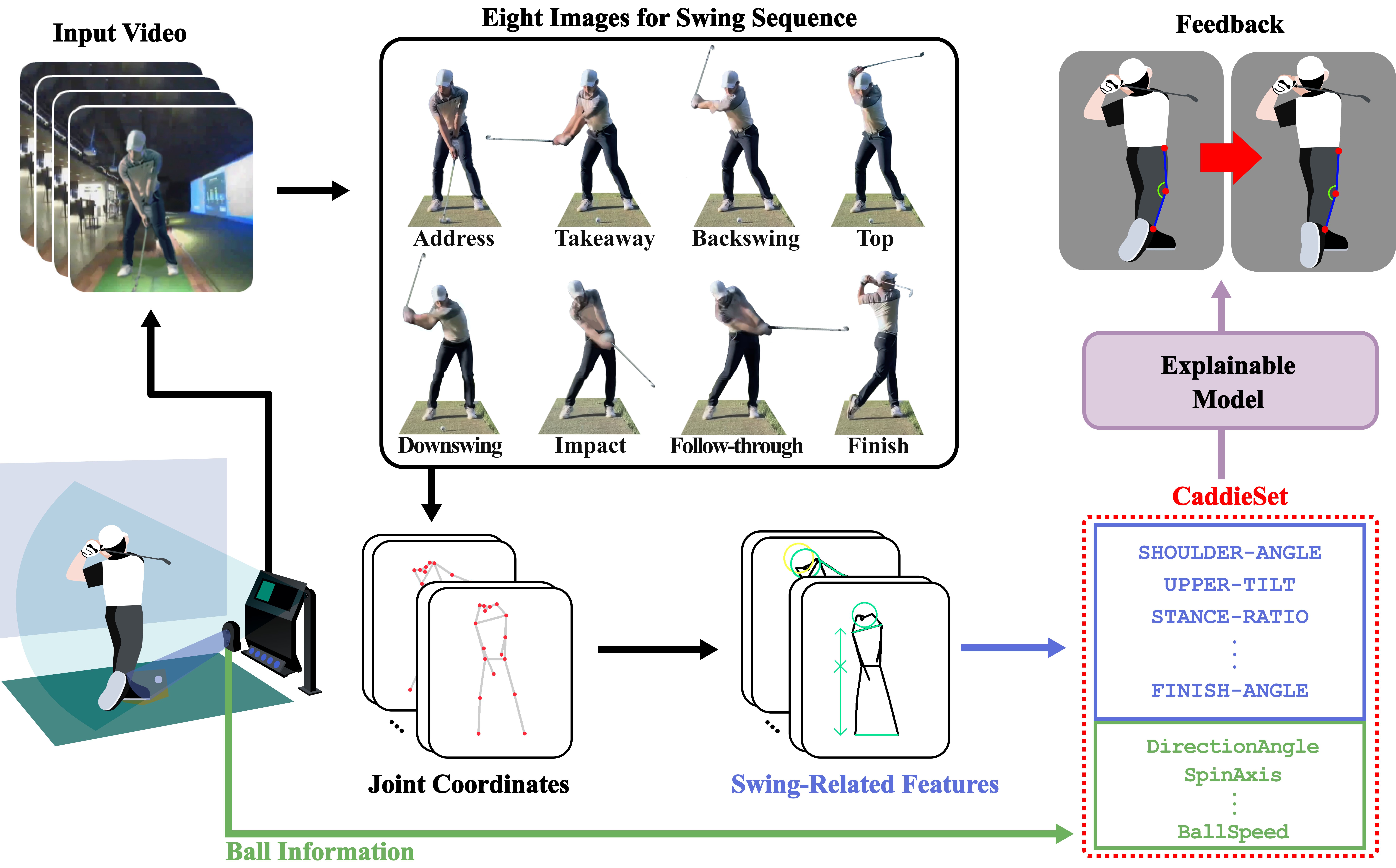}
    \caption{Visualization of our main architecture. Swing video and ball information are imported from the golf launch monitor. Joint information is extracted from the swing video, and joint features are generated based on domain knowledge for use in the experiment. Through this process, the CaddieSet is created, providing a golf swing analysis and individual feedback. In this paper, we focus on using CaddieSet to provide users with feedback on which joint information affects three types of ball information(\textsf{DirectionAngle}, \textsf{SpinAxis}, \textsf{BallSpeed}).}
    \label{fig:fig1}
    \vspace{-2mm}
\end{figure*}

Every golfer aims for a straight shot and adjusts their posture accordingly. However, since a golf swing is a complex movement that involves the entire body, accurately identifying the cause of an off-target shot is challenging \cite{Newell2001the}. In particular, time constraints, high costs, and limited access to personal coaching make it difficult to systematically and continuously refine one's swing.

Recent  Artificial Intelligence (AI) advancements have driven research into AI-based systems for golf swing analysis. For example, some studies have utilized club-attached sensors to track ball trajectories \cite{jiao2018golf, jiao2018multi}, while others have employed computer vision to compare amateur and professional swings to detect incorrect postures \cite{liao2022ai}. However, these methods are limited by sensor dependency or lack of direct analysis of swing and ball trajectory relationships, preventing them from providing precise feedback. 

In particular, existing studies face two major challenges: (1) the lack of detailed analysis utilizing joint information throughout the entire swing sequence, and (2) the absence of paired data linking swing movements with ball trajectory, making it difficult to establish a relationship between them. These limitations hinder the ability to provide accurate and personalized swing correction feedback to golfers.

This study aims to overcome these challenges. In this paper, we propose a system for constructing a new dataset, CaddieSet$^{\dagger}$, which connects golfers' joint information with the corresponding ball trajectory of each shot. Furthermore, we demonstrate the potential of using CaddieSet with interpretable machine learning to provide personalized golf swing corrections.
\footnotetext[2]{https://github.com/damilab/CaddieSet}

The proposed framework of our process is summarized in Figure \ref{fig:fig1}. First, we use a camera-based launch monitor to collect swing videos from various golfers along with corresponding ball trajectory data. Then, using computer vision models, we automatically extract the joint information on the eight events of the golf swing sequence. Next, we generate swing-related features using joint coordinates, guided by domain knowledge. With the help of golf experts, we identify key features derived from swing postures based on these joint values. Through this process, we have identified 15 key features that influence a golf swing, allowing us to analyze the swing's outcome using joint information.

Experimental results demonstrated that the features provided by CaddieSet effectively predict ball trajectory parameters such as ball speed, spin axis, and direction angle. Additionally, these features are interpretable, allowing for precise swing correction and providing detailed feedback for golfers. In quantitative experiments, models trained on joint features, such as Neural Addictive Models (NAM), outperformed image-based models utilizing Vision Transformer (ViT) in predicting ball trajectories. Qualitative experiments further confirmed that explainable models leveraging CaddieSet provided golfers with more intuitive and meaningful insights compared to image-based models.

Finally, we present practical use cases that domain participants can easily understand and apply. Specifically, we demonstrate our swing analysis by comparing the distributions for an individual golfer before and after receiving feedback provided by an explainable machine learning model trained based on CaddieSet. By applying this feedback to adjust the swing posture toward optimal values, we observed significant improvements in swing performance. This approach effectively bridges complex data analysis with practical, actionable insights for golfers. We anticipate that this work will offer a fresh viewpoint on golf swing analysis for both academic research and the sports community.
\section{Related Work}
\label{2}

The golf swing is a sequential chain reaction of the whole body, a series of continuous movements. Although the golf swing is a complex sequence of interconnected actions, it is typically divided into eight events:
\textit{Address}, \textit{Takeaway}, \textit{Backswing}, \textit{Top}, \textit{Downswing}, \textit{Impact}, \textit{Follow-through}, and \textit{Finish} \cite{mcnally2019golfdb}. These steps of swing are shown in Figure \ref{fig:fig1}. Since one action affects all actions performed afterward in the golf swing, correcting wrong postures at each phase can be a fundamental solution to improving golf skills \cite{Newell2001the, zhang2022automatic}.

Extensive research has been conducted on the ideal swing to improve golf performance. Some studies have attached sensors to golfers' bodies to analyze their movements \cite{verikas2017exploring}. For instance, some studies apply computer vision-based pose estimation methods to extract meaningful information from swing videos. Specifically, there have been researches that qualitatively compare the golfers' swing with guideline postures to assist in correcting the swing on their own \cite{liao2022ai, zhang2022automatic}. Because of the limitation of the simple qualitative analyses that compared pose similarities, these studies did not reveal the unique effect of specific body movements on the golf swing performance.

\begin{figure}[htb!]
    \centering
    \includegraphics[width=0.9\linewidth]{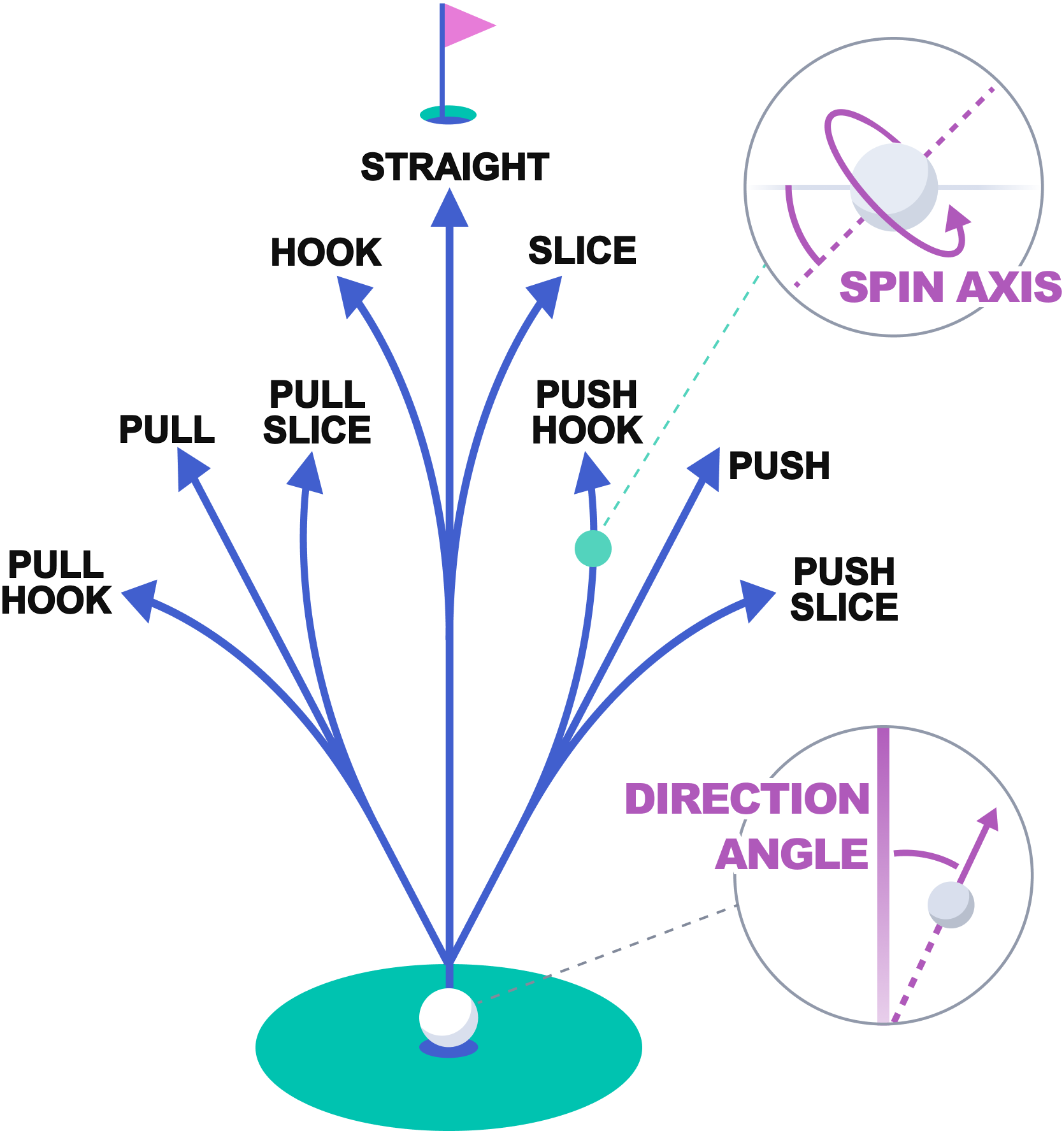}
    \caption{Various trajectories of the golf ball. Depending on the \textsf{DirectionAngle}, the ball will depart left (known as a pull), straight, or right (called a push). Also, the ball trajectory is also influenced by its rotational direction (\textsf{SpinAxis}). A spin causing a leftward curvature is termed a hook, while a rightward curvature is known as a slice.}
    \label{fig:fig2}
    \vspace{-2mm}
\end{figure}

In addition, there has been active research on golf balls, especially focusing on their trajectory. The golf ball flight, known as the trajectory of a golf ball is primarily determined by the point of contact between the club and the ball \cite{sweeney2013influence}. Moreover, to accurately simulate golf ball flight, previous research has suggested a prediction model that combines physical principles and deep learning algorithms \cite{mcnally2023combining}. Two key factors that determine the golf ball's flight are direction angle and spin axis. The direction angle is the initial direction of the ball that starts relative to the target line, and the spin axis represents the amount of curvature of a golf shot. We have visualized the golf ball flight in Figure \ref{fig:fig2}.

Recent studies have explored the relationship between swing and ball trajectory. Some of them have utilized launch monitors such as TrackMan or Skytrak to provide the ball data like flight distance, ball speed, and spin axis \cite{suzuki2021comparison, yamamoto2023extracting}. There have also been attempts to predict the golf ball flight through object detection on the club \cite{yamamoto2023extracting, jiang2022golfpose}. These studies provided insights into both proper swing posture and the correct way to wield the club. Based on \cite{bavcic2016predicting}, they further suggest a meaningful correlation between specific swing planes and golf ball flight. However, obtaining data on factors directly influencing golf ball flight almost invariably requires additional information beyond just body movement or specialized equipment. Consequently, correlating specific joint movements of golfers with their direct impact on golf ball flight remains challenging.

The limitations of previous studies ultimately stem from the lack of datasets that pair corresponding swing information with ball data. GolfDB, a benchmark dataset introduced by \cite{mcnally2019golfdb}, offers high-quality swing information and golfer data. However, it does not include corresponding ball information, limiting its applicability in research. Our CaddieSet provides comprehensive joint features paired with corresponding ball information for each individual's golf swing, effectively addressing this issue.

\section{CaddieSet}
\label{3}

\begin{figure*}[htb!]
    \centering
    \subfloat[Distance Distribution]{
    \includegraphics[width=0.78\textwidth]{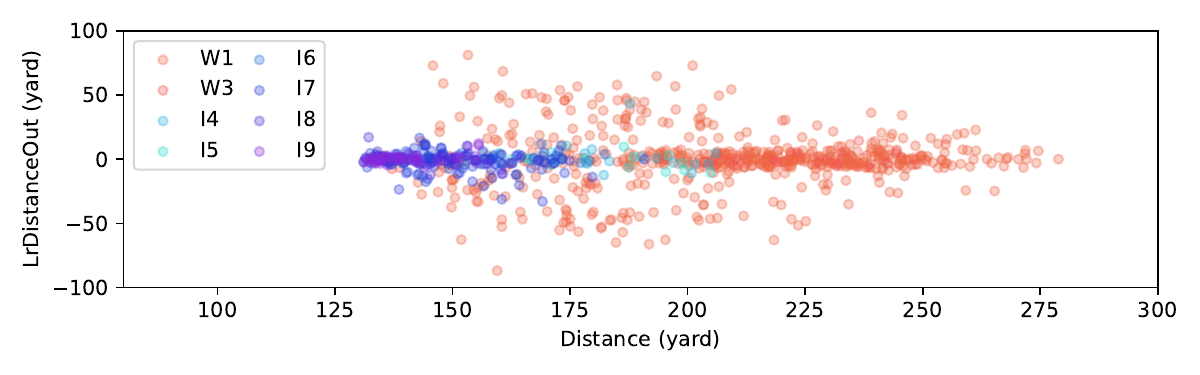}
    \label{fig:fig3_1}}
    \subfloat[Ball Distribution]{
    \includegraphics[width=0.215\textwidth]{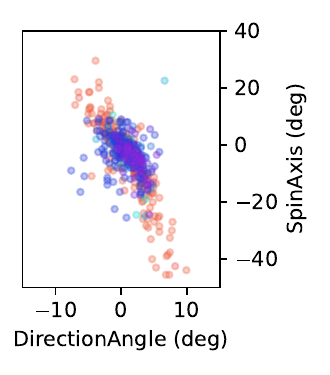}
    \label{fig:fig3_2}}
    \caption{Distribution of ball trajectory data based on \textsf{Distance}, \textsf{LrDistanceOut}, \textsf{DirectionAngle}, and \textsf{SpinAxis}. Different colors represent different golf clubs. Driver is designed to hit the ball the farthest distance, while irons allow for varying shot distances depending on the head angle. The larger the iron number, the shorter the flight distance.}
    \label{fig:fig3}
\end{figure*}

\subsection{Ball Information}
\label{3.1}

Unlike traditional datasets, CaddieSet is a dataset that pairs comprehensive joint features and ball information for each individual's single golf swing. 
Swing videos and ball flight estimates of eight individuals with diverse golf skills were collected using a camera-based launch monitor. Specifically, following the launch monitor's ball measurement methods \cite{kim2013golf, jang2016virtual}, the camera captures the ball trajectory and calculates various ball information.

Each shot was taken from a face-on (\textsf{FACEON}) view and includes information on club type (\textsf{ClubType}), flight distance (\textsf{Distance}, \textsf{Carry}), left and right distance (\textsf{LrDistanceOut}), direction angle (\textsf{DirectionAngle}), spin axis (\textsf{SpinAxis}), and ball speed (\textsf{BallSpeed}). The \textsf{ClubType} used include driver (W1), wood (W3), and irons (I4, I5, I6, I7, I8, I9). The \textsf{DirectionAngle} represents the angle at which the ball deviates from the target line, and the \textsf{SpinAxis} describes the tilt of the ball's spin, which affects its trajectory and curvature. The \textsf{BallSpeed} refers to the velocity of the ball in flight. CaddieSet comprises 924 shots, including 613 with driver (W1), 23 with 3 wood (W3), 288 with irons (from 4 to 9, I4-I9).

The ball information in CaddieSet is shown in Figure \ref{fig:fig3}. Figure \ref{fig:fig3_1} represents the landing point of the ball for one shot, the x-axis represents the \textsf{Distance}, and the y-axis represents the \textsf{LrDistanceOut} from the target. When club length increases (from I9 to W1), flight distance increases, and left and right distance also increase. Figure \ref{fig:fig3_2} shows the \textsf{DirectionAngle} and \textsf{SpinAxis} for each shot. A negative \textsf{DirectionAngle} means the ball starts left, while a positive value means it starts right. A negative \textsf{SpinAxis} indicates clockwise spin, and a positive value indicates counterclockwise spin. Figure \ref{fig:fig3_2} explains the various golf ball trajectories shown in Figure \ref{fig:fig2}. For example, the shot in the upper left corner is a `pull slice' shot that starts toward the left but curves to the right. These figures demonstrate that CaddieSet encompasses a variety of shot types, derived from a large number of swing videos.

\subsection{Extracting Joint Coordinates}
\label{3.2}

Extracting corresponding frames from the swing video is necessary to obtain human postures of eight golf swing events. We used the fine-tuned SwingNet architecture for the golf swing sequencing network. The training dataset comprised 1,050 videos from GolfDB \cite{mcnally2019golfdb} and other swing videos, which include various view types of golfers with diverse skill levels. For validation, we used 350 videos from part of GolfDB. During training, horizontal flipping and affine transformation were used for data augmentation, and other training configurations followed those in \cite{mcnally2019golfdb}.

\begin{table*}[htb!]
    \centering
    \caption{Description of the 15 metrics: Each indicator measures specific joint information in the golf swing. These metrics help evaluate and improve the efficiency and effectiveness of the swing. They are based on insights from domain experts.}
    \resizebox{0.9\linewidth}{!}{%
    \begin{tabular}{lll}
        \hline
        \textbf{Metric}               & \textbf{Description}                                                           & \textbf{Measurement} \\ \hline
        \texttt{SHOULDER-ANGLE}        & Shoulder angle relative to horizontal                                          & degree   \\
        \texttt{UPPER-TILT}            & Lower body to upper body ratio                                                 & ratio    \\
        \texttt{STANCE-RATIO}          & Shoulder width to stride length ratio                                          & ratio    \\
        \texttt{HEAD-LOC}              & Movement of head relative to \textit{Address}                                  & ratio    \\
        \texttt{SHOULDER-LOC}          & Left shoulder position within the width of the stride                          & ratio    \\
        \texttt{LEFT-ARM-ANGLE}        & Angle formed by left shoulder, elbow, and wrist                                & degree   \\
        \texttt{RIGHT-ARM-ANGLE}       & Angle formed by right shoulder, elbow, and wrist                               & degree   \\
        \texttt{HIP-ROTATION}          & Rotation degree of pelvis relative to \textit{Address}                         & degree   \\
        \texttt{HIP-SHIFTED}           & Movement of hip relative to \textit{Address}                                   & ratio    \\
        \texttt{RIGHT-LEG-ANGLE}       & Angle formed by right hip, knee, and ankle                                     & degree   \\
        \texttt{SHOULDER-HANGING-BACK} & Relative distance between left ankle and left shoulder to stride length ratio  & ratio    \\
        \texttt{HIP-HANGING-BACK}      & Relative distance between left ankle and left hip to stride length ratio       & ratio    \\
        \texttt{RIGHT-ARMPIT-ANGLE}    & Angle formed by right elbow, shoulder, and pelvis                              & degree   \\
        \texttt{WEIGHT-SHIFT}          & Angle formed by a line connecting left ankle to left pelvis                    & degree   \\
        \texttt{FINISH-ANGLE}          & Angle formed by a line connecting left ankle to right hip                      & degree   \\ \hline
    \end{tabular}%
    }
    \label{tab:table1}
\end{table*}

\begin{figure*}[htb!]
    \centering
    \includegraphics[width=0.8\linewidth]{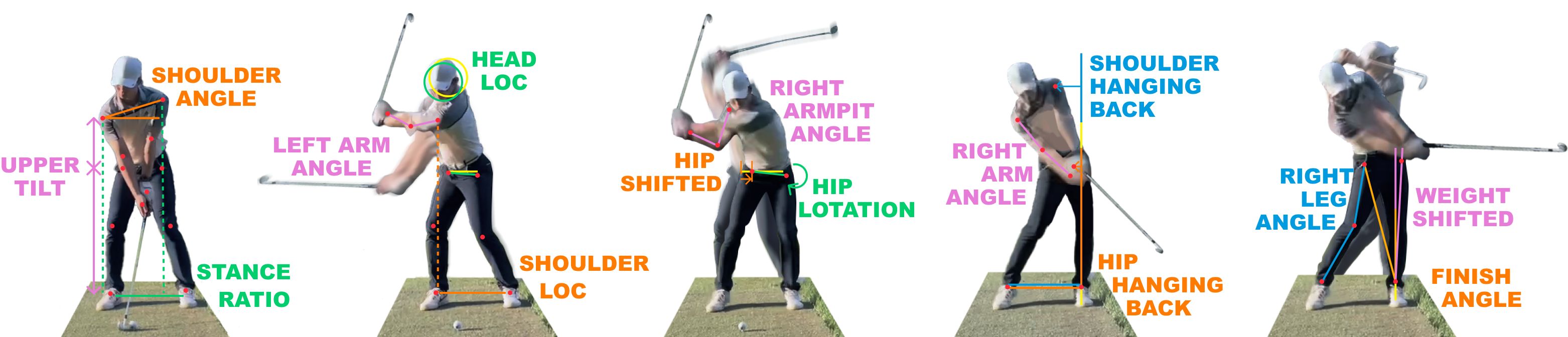}
    \caption{15 types of swing-related joint information generated from a \textsf{FACEON} view golf swing video. The golf swing is classified into eight swing events, and the red dot represents the extracted joint location. Additionally, we extracted swing-related joint changes using yellow lines for the original position and green lines for the adjusted position.}
    \label{fig:fig4}
    \vspace{-2mm}
\end{figure*}

Once the sequence mapping model generates eight swing sequential images, each image is processed through Human Detection and Human Pose Estimation models to predict the bounding box and joint coordinates of the golfer, respectively. We opted to employ widely used architectures, Faster R-CNN \cite{ren2015faster} for object detection and HRNet \cite{sun2019deep} for pose estimation. The bounding box predictions from Faster R-CNN and the swing sequence image are passed through HRNet to yield 17 human joint coordinates (`Nose', `L\_Eye', `R\_Eye', `L\_Ear', `R\_Ear', `L\_Shoulder', `R\_Shoulder', `L\_Elbow', `R\_Elbow', `L\_Wrist', `R\_Wrist', `L\_Hip', `R\_Hip', `L\_Knee', `R\_Knee', `L\_Ankle', `R\_Ankle'). For detailed information and performance evaluation on the models used, refer to the supplementary material \ref{S.A}.

\subsection{Generating Swing-Related Features} 
\label{3.3}

The posture components affecting the ball flight are estimated for each swing posture based on the obtained joint values. All joint coordinates are scaled with the bounding box width, considering different camera positions during data acquisition. 15 analysis metrics were established based upon \cite{mann1998swing, wiren1991golf}, to utilize professional golf domain knowledge. The method for calculating each metric is as follows Table \ref{tab:table1}, and all 15 metrics are illustrated in Figure \ref{fig:fig4}.

All 15 metrics play a crucial role in determining how far and accurately the ball flights. For example, metrics such as \texttt{SHOULDER-ANGLE}, \texttt{HIP-ROTATION}, \texttt{WEIGHT-SHIFT}, \texttt{RIGHT-LEG-ANGLE} reflect the importance of maintaining body balance and efficiently transferring power until impact. Also, metrics like \texttt{HEAD-LOC}, \texttt{SHOULDER-LOC}, \texttt{STANCE-RATIO} indicate the importance of maintaining the correct swing plane and consistency for accurate ball striking. By analyzing and improving these metrics, golfers can enhance both the power and precision of their swings. The metrics are measured using either degrees or ratios. While degrees provide a straightforward angle measurement, ratios are used to represent relative positions or changes in position during continuous sequences. For example, in the case of \texttt{HEAD-LOC}, the ratio indicates the change in the position of the head relative to the \textit{Address} position.

\begin{figure*}[htb!]
    \centering
    \includegraphics[width=0.8\linewidth]{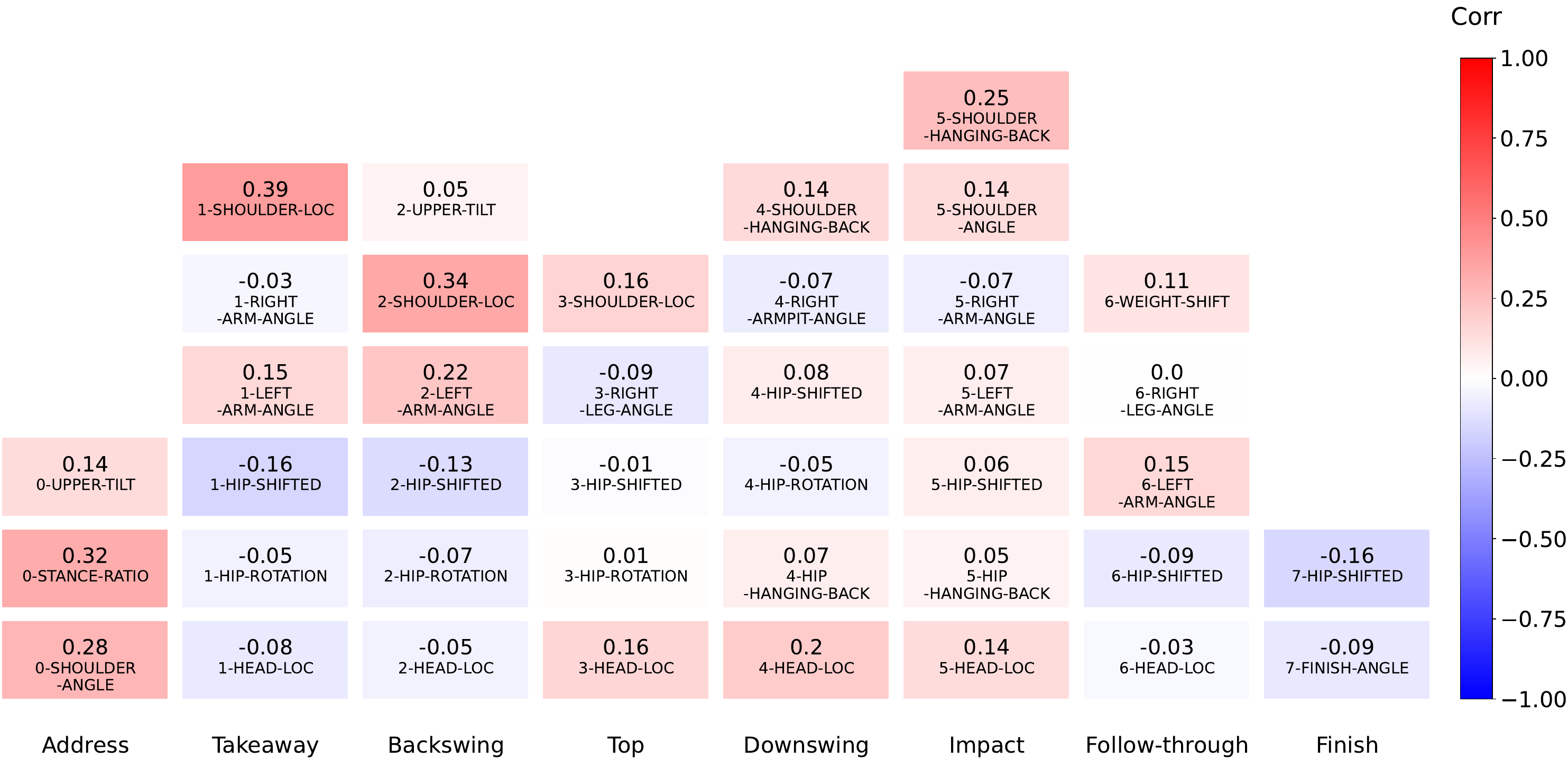}
    \caption{Correlation heatmap between swing-related features and \textsf{BallSpeed}. In the case of \texttt{STANCE-RATIO}, there is a significant positive correlation of 0.32. \texttt{STANCE-RATIO}, which refers to the ratio of shoulder width to foot stride length, allows for effective power and weight transfer, increasing ball speed and distance. Additionally, shoulder-related features such as \texttt{SHOULDER-LOC} and \texttt{SHOULDER-ANGLE} show a higher correlation than other features, indicating their critical role in enhancing swing efficiency and power.}
    \label{fig:fig5}
    \vspace{-2mm}
\end{figure*}

Based on domain knowledge, 15 metrics were used as important values for each of the eight swing events. These 40 swing-related features, along with their utilization at each stage, are depicted in Figure \ref{fig:fig5}. For ease of identifying variables in any swing event and distinguishing redundant variables across multiple stages, each feature is assigned a number from 0 to 7 corresponding to the swing events. For the rationale behind swing metric assignments and their roles in each phase, refer to the supplementary material \ref{S.B}.

We conducted a detailed analysis of the correlation between all swing-related features and golf ball speed to confirm their relevance, as shown in Figure \ref{fig:fig5}. The features we investigated showed a significant association with ball trajectory. For instance, the \textit{Address} phase features like \texttt{STANCE-RATIO} and \texttt{SHOULDER-ANGLE} are correlated with swing efficiency and power transfer, leading to improvements in \textsf{BallSpeed}.
\section{Benchmark Evaluation}
\label{4}

To verify the effectiveness of CaddieSet across various models and tasks, we conducted experiments using multiple benchmarks. We employed eight different methods to explore how well CaddieSet performs. Firstly, we employ vision-based models such as ResNet \cite{he2016deep}, MobileNet \cite{howard2017mobilenets}, and Vision Transformer (ViT) \cite{dosovitskiy2020image} to process input images directly from golf swing videos. The details on the experimental setup are summarized in the supplementary material \ref{S.C}. In addition, we used four traditional machine learning methods that leverage the swing-related features from CaddieSet: Linear/Logistic Regression (LR), Support Vector Machine (SVM), Random Forest (RF), and XGBoost. Furthermore, to provide intrinsic explanations through CaddieSet, we also trained Neural Additive Models (NAM) \cite{agarwal2021neural, kim2024bayesnam}, which was recently proposed as an explainable AI (XAI). This feature-based method offers greater interpretability than image-based models, providing users with more understandable explanations of the results.

The target variables used in the analysis are \textsf{DirectionAngle}, \textsf{SpinAxis}, and \textsf{BallSpeed}. Since hitting the ball straight is crucial, as shots that deviate to the left or right are undesirable, \textsf{DirectionAngle} and \textsf{SpinAxis} were processed as binary classifications, aiming to clearly distinguish between straight and non-straight shots. These classifications were evaluated using accuracy (Acc) and Area Under the ROC Curve (AUC) as the evaluation metrics. Specifically, for the \textsf{DirectionAngle}, values within $\pm$6 degrees from the straight path were classified as straight, while those beyond this threshold were considered non-straight. Similarly, for the \textsf{SpinAxis}, a threshold of $\pm$10 degrees was used to differentiate between desirable and undesirable spin characteristics. Additionally, sending the ball a long distance requires achieving higher velocity. To accomplish this, \textsf{BallSpeed} was analyzed using regression to predict the continuous variation in ball speed based on swing-related features, with mean squared error (MSE) as the evaluation metric. Of the 924 data from CaddieSet, 739 samples were used for training and 185 for testing. The experimental results are summarized in Table \ref{tab:table2}.

\begin{table}[htb!]
    \centering
    \caption{Benchmark comparison of model performance on the target variables \textsf{DirectionAngle}, \textsf{SpinAxis}, and \textsf{BallSpeed}. The best experimental results are in bold.}
    \resizebox{0.99\columnwidth}{!}{%
    \begin{tabular}{cccccc}
    \hline
    \multirow{2}{*}{\textbf{Method}} & \multicolumn{2}{c}{\textsf{DirectionAngle}} & \multicolumn{2}{c}{\textsf{SpinAxis}} & \textsf{BallSpeed}    \\ \cline{2-6} 
                                     & \textbf{Acc}     & \textbf{AUC}    & \textbf{Acc}  & \textbf{AUC} & \textbf{MSE} \\ \hline
    \textbf{ResNet18}                & \textbf{0.9027}        & 0.8217                & \textbf{0.7297}     & 0.7431             & 49.90             \\
    \textbf{MobileNet\_V3}           & 0.8973                 & 0.8510                & 0.7081              & 0.7743             & 32.32             \\
    \textbf{ViT-B/16}                & 0.6270                 & 0.7005                & 0.6865              & 0.6738             & 48.77             \\ \hline
    \textbf{LR}                      & 0.8973                 & 0.7811                & 0.6703              & 0.7065             & 10.82             \\
    \textbf{SVM}                     & 0.8595                 & 0.7027                & 0.6757              & 0.7322             & 38.83             \\
    \textbf{RF}                      & 0.8973                 & 0.8004                & 0.7027              & 0.7551             & \textbf{8.80}     \\
    \textbf{XGBoost}                 & 0.8865                 & 0.7929                & 0.6865              & 0.7379             & 10.06             \\
    \textbf{NAM}                     & 0.8162                 & \textbf{0.8757}       & 0.6811              & \textbf{0.7851}    & 9.72             \\ \hline
    \end{tabular}%
    }
    \label{tab:table2}
\end{table}

When constructing swing-related features from the original swing image data, there is a potential risk of information loss. Despite this, our evaluation demonstrates that even with the derived features alone, substantial performance can be achieved. This is evidenced by the competitive results of our feature-based models. By validating that the swing-related features can perform comparably to the original data, we affirm that CaddieSet is a valuable resource for analyzing and improving golf swings.

\begin{figure*}[htb!]
    \centering
    \begin{subfigure}[b]{0.3\textwidth}
        \centering
        \includegraphics[width=\linewidth]{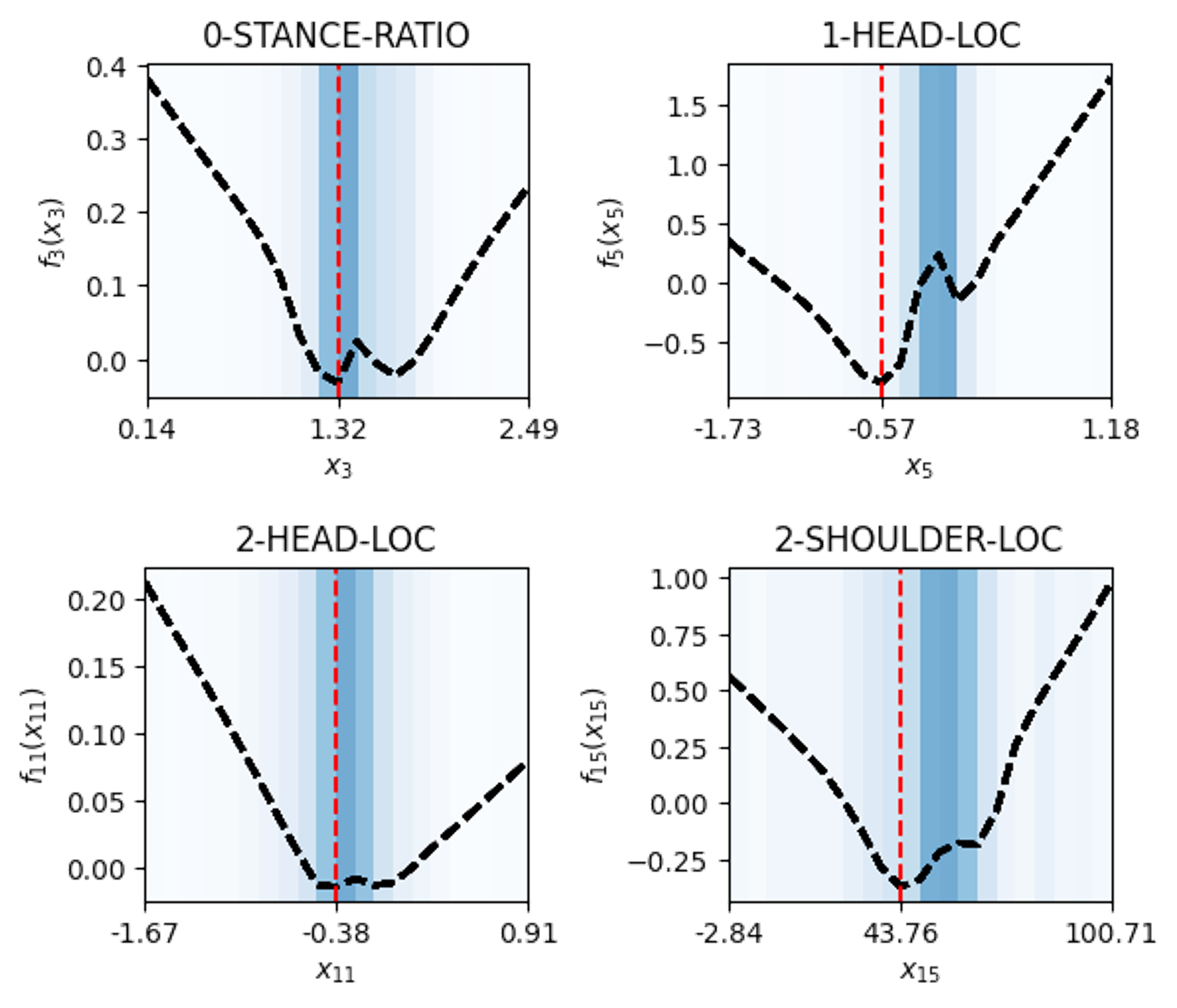}
        \caption{\textsf{DirectionAngle}}
        \label{fig:fig8_1}
    \end{subfigure}
    \hfill
    \begin{subfigure}[b]{0.3\textwidth}
        \centering
        \includegraphics[width=\linewidth]{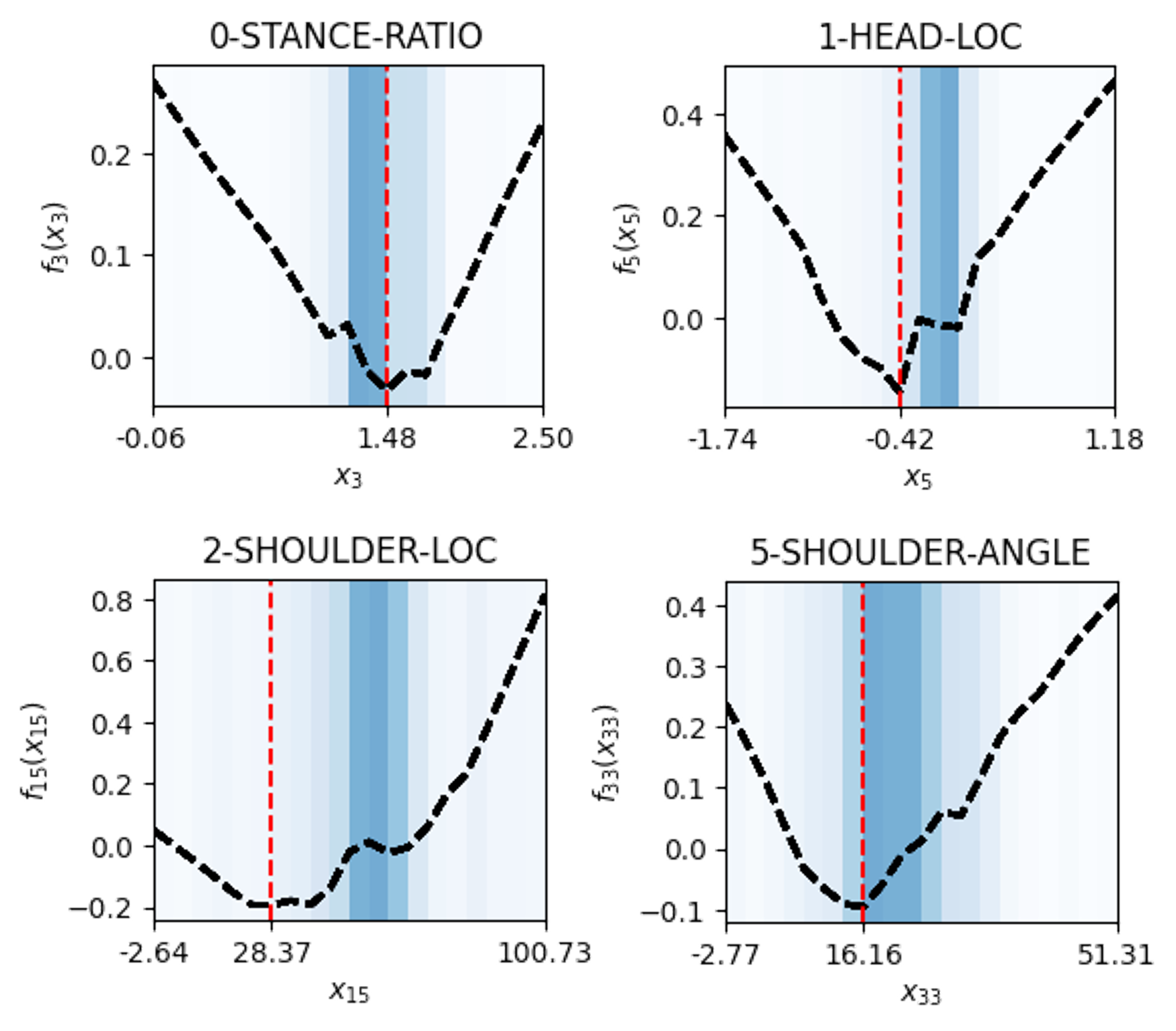}
        \caption{\textsf{SpinAxis}}
        \label{fig:fig8_2}
    \end{subfigure}
    \hfill
    \begin{subfigure}[b]{0.3\textwidth}
        \centering
        \includegraphics[width=\linewidth]{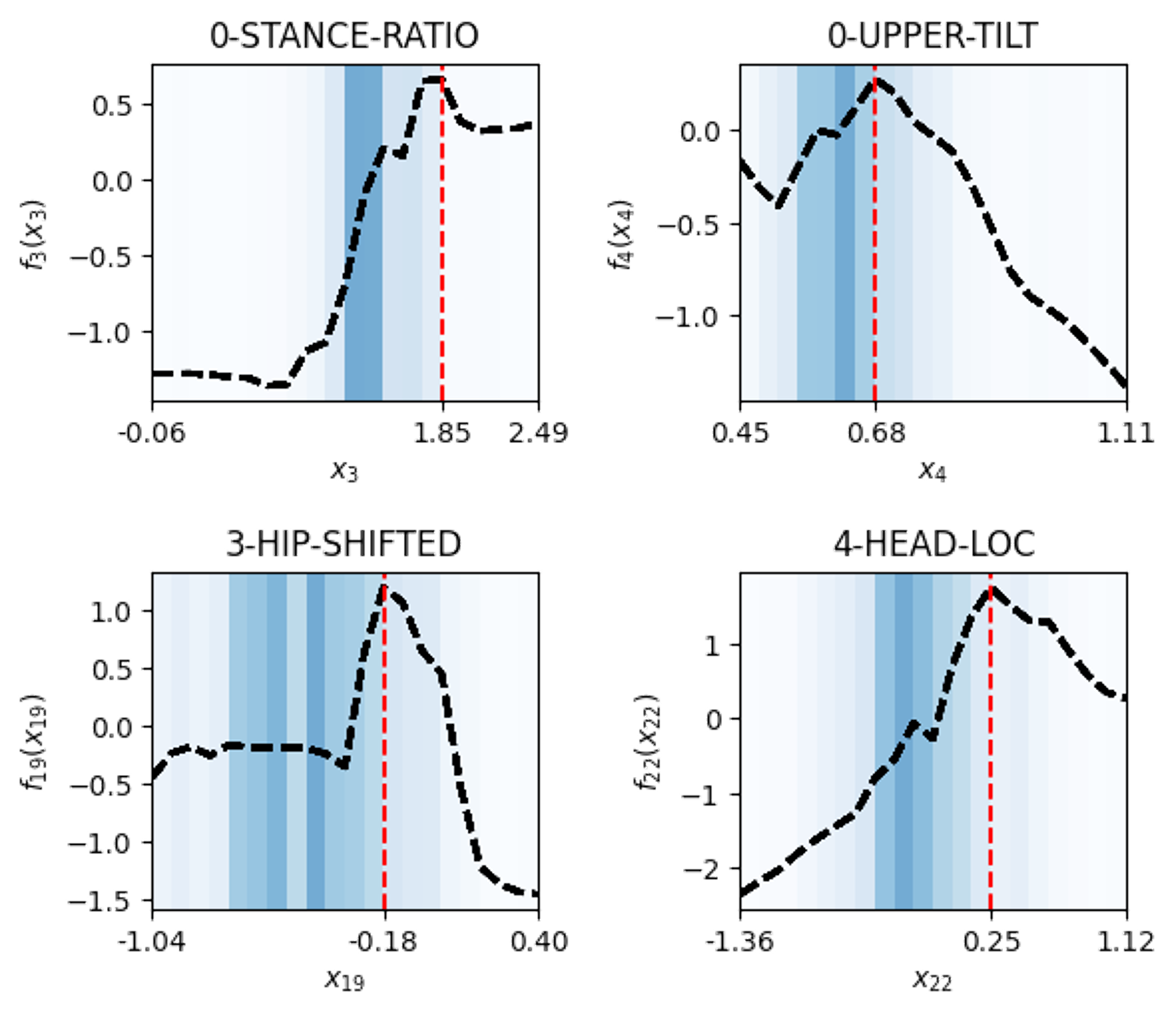}
        \caption{\textsf{BallSpeed}}
        \label{fig:fig8_3}
    \end{subfigure}
    \caption{Graphs of swing-related features and ball trajectory. The x-axis of the graph is the value of the feature, and the y-axis represents the effect that feature value has on the prediction. The black dotted line indicates the function shape and the red line highlights the value of $x_i$ when the outstanding value of $f_i(x_i)$. The functional shape $f_i(x_i)$ of the important explanatory variable $x_i$ and the distribution of $x_i$ in the high-density region. The background color corresponds to the distribution of swing-related features, with darker colors indicating higher density.}
    \label{fig:fig8}
    \vspace{-2mm}
\end{figure*}

\begin{figure}[htb!]
    \centering
    \includegraphics[width=0.9\linewidth]{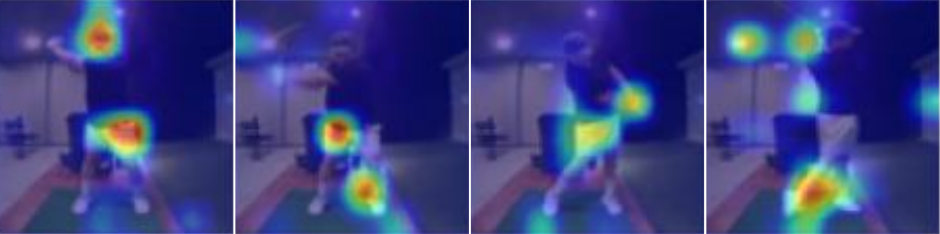}
    \caption{The attention map of the ViT model targeting \textsf{BallSpeed}. This map highlights and emphasizes the importance of the golfer's posture, shoulder rotation, and overall club movement throughout the swing.}
    \label{fig:fig7}
    \vspace{-2mm}
\end{figure}

\section{Application for Golf Performance}
\label{5}

\subsection{Explainable Methods for Swing Analysis}
\label{5.1}

In the realm of sports, the advent of AI has equipped teams and athletes with powerful tools to analyze vast amounts of data for competitive advantage. However, as AI becomes more prevalent in sports analytics, one critical aspect emerges: \textit{explainability}. Understanding how AI models arrive at their predictions is crucial for athletes and coaches to trust and effectively utilize these insights.

In this section, we focus on comparing two models among eight benchmarks in the previous section: ViT, an image-based model, and NAM, a feature-based model. ViT is a powerful computer vision method that uses attention maps to identify which parts of a golf swing the model focuses on. As shown in Figure \ref{fig:fig7}, attention maps of ViT can qualitatively show where the model is concentrating. However, they cannot quantitatively analyze how much each part (e.g., joint) affects the outcome. This limitation makes it challenging to provide golfers with direct, actionable feedback, as it is difficult to measure the precise impact of each feature on the swing's results.

Unlike ViT, NAM offers quantitative insights into the influence of each feature on the outcome, making it easier to provide precise feedback to users. NAM describes the relationship as a sum of functions, each representing the impact of individual input features:
\begin{equation}\label{eq:nam}
    y = f_1(x_1) + f_2(x_2) + \cdots + f_d(x_d) + \beta,
\end{equation}
This clarity improves the model's usability and helps identify critical elements that contribute to performance improvements, providing direct feedback to golfers.

Our model can influence the golf swing based on various joint information, as previously described in Section \ref{3.3}. Additionally, as explained in Section \ref{3.1}, it contains various ball information depending on the swing, such as the \textsf{DirectionAngle}, \textsf{SpinAxis}, and \textsf{BallSpeed}. For example, the second graph of Figure \ref{fig:fig8_2} explains the nonlinear relationship between \texttt{2-HEAD-LOC} and \textsf{SpinAxis}. Therefore, our framework allows the users to analyze which values of various features affect the target and makes it easy to interpret non-linear relationships.

We interpret the independent impact of each feature on the final prediction using the output graph of each neural network $f_i$  as described in Equation \ref{eq:nam}. In Figure \ref{fig:fig8}, the x-axis represents the value of the feature, while the y-axis indicates the influence of that feature on the model's prediction. For \textsf{DirectionAngle} and \textsf{SpinAxis}, feature values that align with optimal swing mechanics (such as a balanced stance or ideal shoulder position) increase the likelihood of achieving a straight shot, indicating a more desirable swing outcome. Conversely, deviations in these feature values from the optimal range can lead to non-straight shots, reflecting a less favorable swing outcome. For Figure \ref{fig:fig8_3}, the sum of $f_i(x_i)$ represents the overall \textsf{BallSpeed}. Therefore, a higher $f_i(x_i)$ value for a particular $x_i$ indicates a positive effect on the ball speed, suggesting that optimizing these features can enhance performance.

\vspace{-5mm}
\paragraph{Minimizing \textsf{DirectionAngle} requires a stable head position and precise left shoulder alignment, ensuring proper swing mechanics and trajectory.} The \texttt{0-STANCE-RATIO} measures the stance width relative to shoulder width, critical for balance and directional control. A stance that is either too narrow or too wide can disrupt the swing's stability, leading to reduced accuracy and consistency. Figure \ref{fig:fig8_1} indicates an optimal range around 1.32 where the effect on direction is minimized, suggesting that maintaining a balanced stance is key to reducing directional deviation. Stability in head position is essential for maintaining a consistent swing plane. The graphs show a minimal effect on \textsf{DirectionAngle} when the head position is near -0.56 (\texttt{1-HEAD-LOC}) and -0.38 (\texttt{2-HEAD-LOC}). This highlights the importance of a stable head position to ensure the ball travels straight, providing insight into addressing head movement issues, such as head-up errors during the swing. Proper shoulder alignment and rotation are crucial for ensuring the swing path aligns correctly with the intended direction. \texttt{2-SHOULDER-LOC} shows that deviations from optimal shoulder alignment can significantly influence the \textsf{DirectionAngle}, underscoring the importance of correct shoulder rotation during the \textit{Backswing} to ensure accurate ball contact.

\vspace{-5mm}
\paragraph{Achieving an optimal \textsf{SpinAxis} requires precise shoulder alignment and control over the stance, which are essential for minimizing unnecessary ball curvature and ensuring accuracy.} Similar to \textsf{DirectionAngle}, a balanced \texttt{0-STANCE-RATIO}, stable head position \texttt{1-HEAD-LOC} and proper shoulder alignment \texttt{2-SHOULDER-LOC} are crucial for controlling spin and ensuring the ball travels accurately. These factors play a significant role in minimizing side spin, thus enhancing shot precision. Moreover, the \texttt{5-SHOULDER-ANGLE} in Figure \ref{fig:fig8_2} highlights that maintaining a shoulder angle of around 16.16 degrees is optimal for minimizing spin axis deviation. This aligns with traditional golf knowledge that suggests shoulders should be slightly open at impact to square the clubface with the target, improving directional control and reducing spin-induced ball curvature.

\vspace{-5mm}
\paragraph{To increase \textsf{BallSpeed}, maintaining a balanced stance and optimizing upper body tilt are essential for effective power transfer and shot distance.} As Figure \ref{fig:fig8_3}, there is a sharp increase in \textsf{BallSpeed} when the \texttt{0-STANCE-RATIO} is around 1.85. This suggests that a wider stance within this range allows for better stability and power generation during the swing. The \texttt{0-UPPER-TILT} highlights the importance of upper body positioning. The graph indicates that maintaining an upper body tilt of around 0.68 maximizes leverage and force, leading to increased \textsf{BallSpeed}. The metric emphasizes the role of hip movement. This suggests that an effective hip shift and proper rotation during the \textit{Backswing} and \textit{Downswing} are crucial for generating torque and power, enabling efficient energy transfer to the ball. Lastly, the \texttt{4-HEAD-LOC} demonstrates the impact of head positioning on \textsf{BallSpeed}. The optimal range around 0.25 in Figure \ref{fig:fig8_3} indicates that keeping the head steady and within this range can help maintain a consistent swing path and maximize energy transfer to the ball.

\subsection{Practical Examples of Individual Feedback}
\label{5.2}
In this section, we experimentally analyze whether the swing analysis provided by the XAI model trained with the CaddieSet proposed in Section \ref{5.1} is effective in correcting the swings of actual golfers.

We demonstrated a specific case of providing detailed feedback to a single amateur golfer. The golfer completed 10 swings before receiving feedback. Next,  by using NAM model trained on CaddieSet for swing analysis, we obtained results similar to those shown in Figure \ref{fig:fig8}, which allowed us to provide quantitative feedback on the golfer's swing. After providing feedback, the golfer performed another 10 swings, to analyse the effective of feedback given by NAM.

\begin{figure}[htb!]
    \centering
    \begin{subfigure}{0.43\columnwidth}
        \centering
        \includegraphics[width=\linewidth]{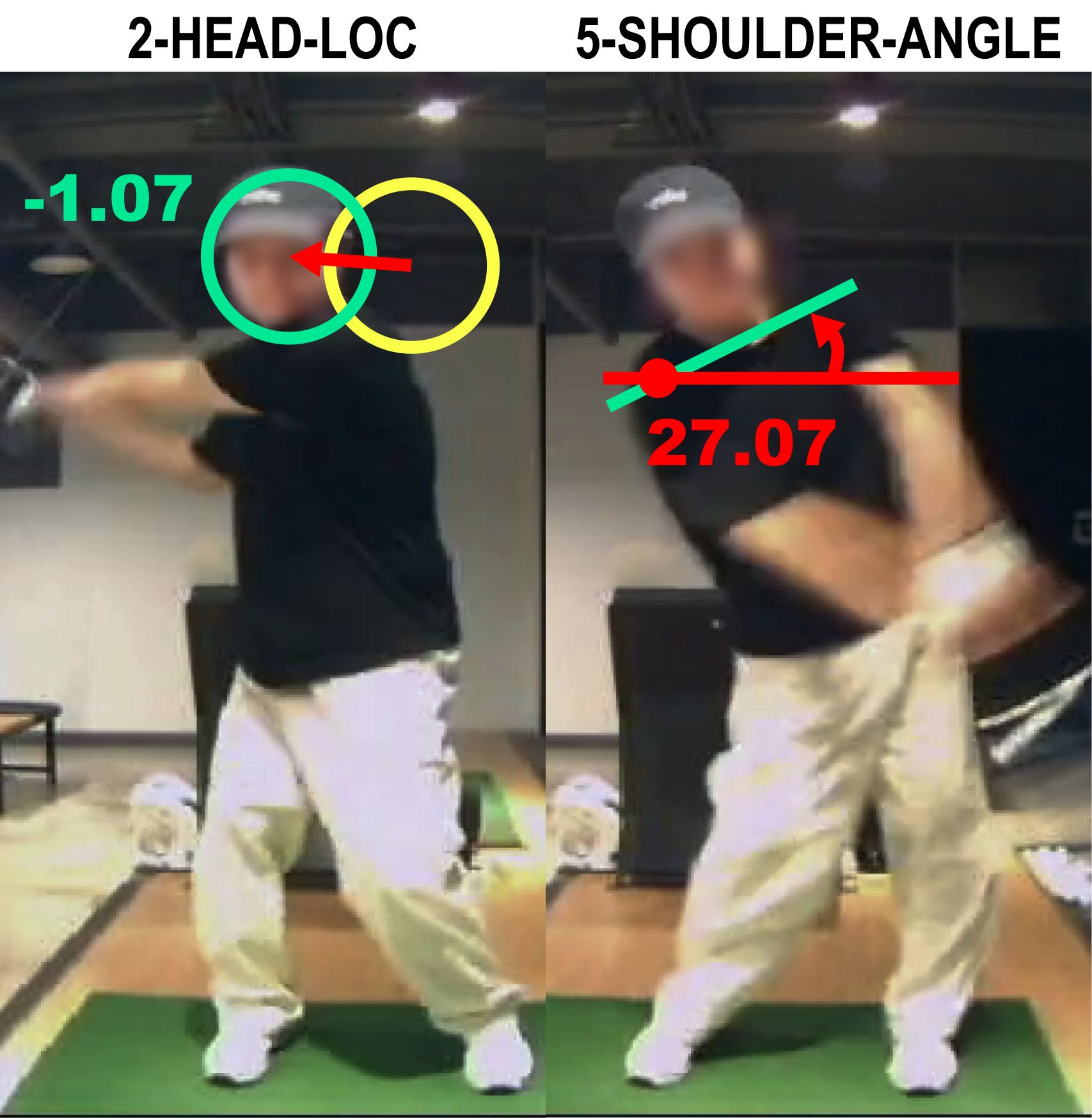}
        \caption{\textsf{Before}}
        \label{fig:fig10_1}
    \end{subfigure}
    \hfill
    \begin{subfigure}{0.43\columnwidth}
        \centering
        \includegraphics[width=\linewidth]{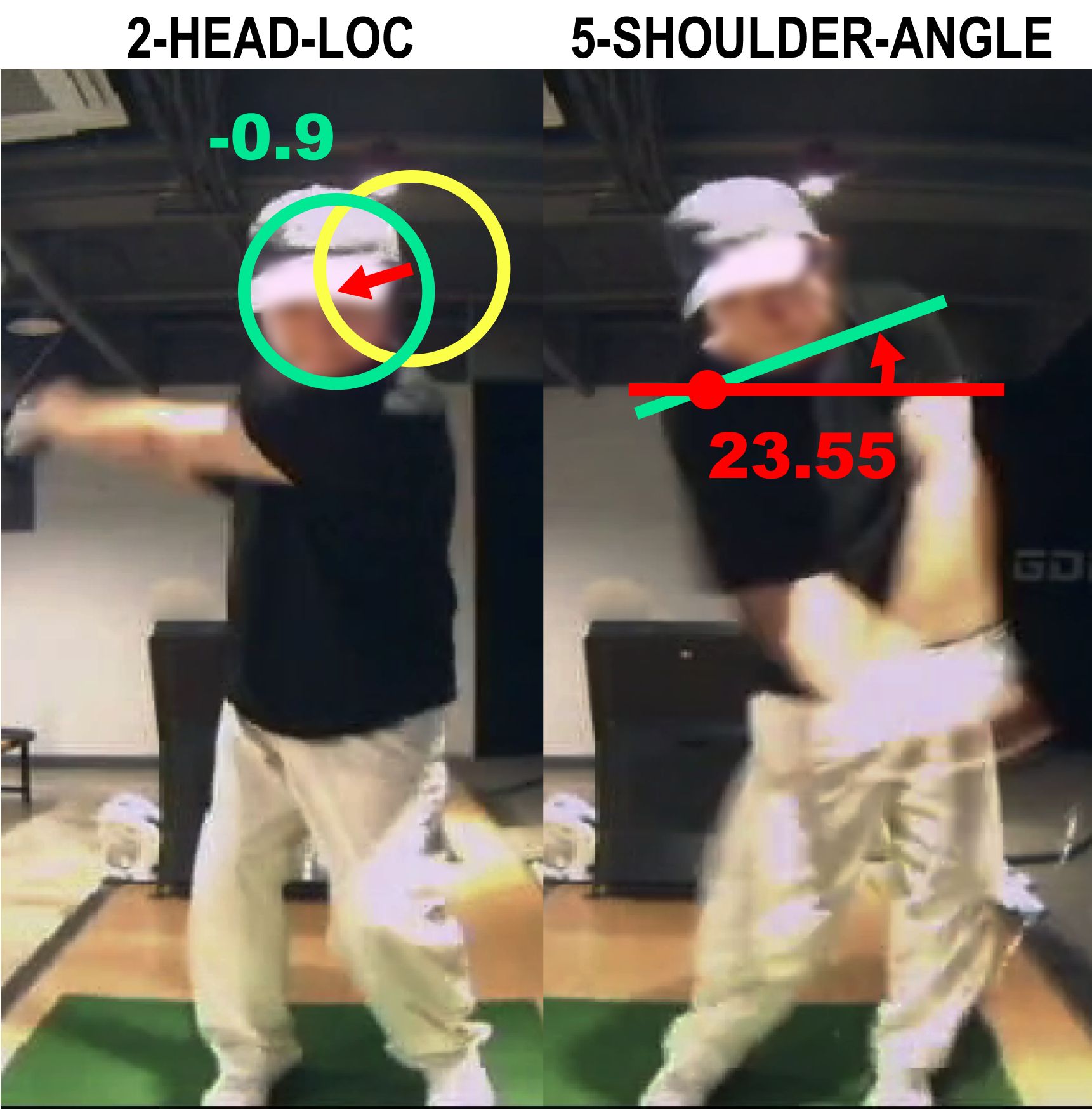}
        \caption{\textsf{After}}
        \label{fig:fig10_2}
    \end{subfigure}
    \caption{Comparison of the golfer's posture before and after feedback. Based on the explanation drawn from the model, we provided feedback to the golfer on \texttt{2-HEAD-LOC} and \texttt{5-SHOULDER-ANGLE}, helping to correct their posture.}
    \label{fig:fig10}
    \vspace{-2mm}
\end{figure}

To improve the \textsf{DirectionAngle}, it is important to adjust the golfer's \texttt{2-HEAD-LOC}. Before receiving feedback, the golfer's average \texttt{2-HEAD-LOC} was -1.07, as shown in Figure \ref{fig:fig10_1}. Figure \ref{fig:fig8_1} shows that the ideal value for \texttt{2-HEAD-LOC} to straighten the angle is -0.38. We provided feedback aimed at adjusting their \texttt{HEAD-LOC} closer to ideal value. After the feedback, as shown as Figure \ref{fig:fig10_2}, the golfer's \texttt{2-HEAD-LOC} average has moved to -0.9. We also observed improvements in \textsf{SpinAxis} through feedback. To achieve the ideal values for \textsf{SpinAxis}, making the ball travel straighten and to not curve, we provided feedback to adjust the \texttt{5-SHOULDER-ANGLE} closer to 16.16. In detail, the golfer's \texttt{5-SHOULDER-ANGLE} mean has changed to 23.55 from 27.07.
\begin{figure}[htb!]
    \centering
    \includegraphics[width=0.8\linewidth]{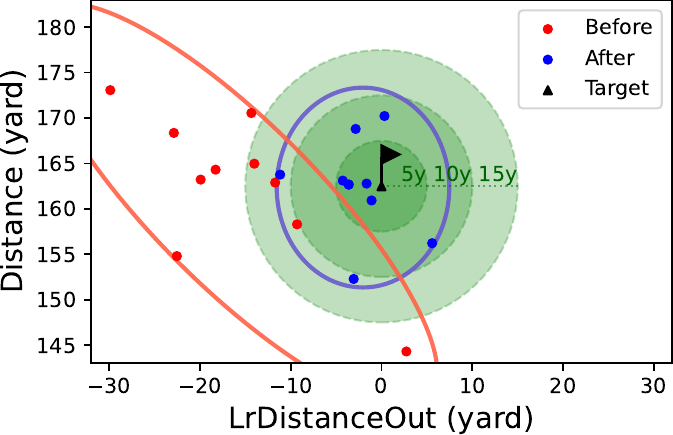}
    \caption{Ball lie distribution before and after feedback. The Target is positioned with an \textsf{LrDistanceOut} of zero, and the \textsf{Distance} is at the average for the golfer. Each red and blue dot represents the lie, where the dropped ball comes to rest, before and after receiving feedback, respectively.}
    \label{fig:fig11}
    \vspace{-2mm}
\end{figure}

Following our feedback, we observed improvements in the golfer's posture during their swing, leading to a more controlled and predictable ball trajectory. As shown in Figure \ref{fig:fig11}, the shot distribution became more consistent, demonstrating that our feedback helped correct the golfer's posture and resulted in improved shots. By applying this feedback, we confirmed the significance of the ideal joint positions identified in our analysis, effectively enhancing the ball trajectory. This result shows the effectiveness of the swing analysis and feedback provided by the model trained on CaddieSet.
\section{Conclusion}

In this paper, we introduced CaddieSet, a novel dataset for golf swing analysis. We extracts human joint features from videos and pairs them with corresponding ball information for each swing. Through various benchmarks, we validated the effectiveness of the dataset, demonstrating that the features in CaddieSet can be utilized effectively for golf swing analysis. Our experiments showcased the practical application of CaddieSet for swing analysis and personalized feedback for individual golfers. 
\paragraph{Acknowledgment} This work was supported in part by Kimcaddie Inc, and the MSIT(Ministry of Science and ICT), Korea, under the ITRC(Information Technology Research Center) support program(IITP-2025-2020-0-01789), and the Artificial Intelligence Convergence Innovation Human Resources Development(IITP-2025-RS-2023-00254592) supervised by the IITP(Institute for Information \& Communications Technology Planning \& Evaluation), and the National Research Foundation of Korea(NRF) grant funded by the Korea government(MSIT) (RS-2025-02219688).
{
    \small
    \bibliographystyle{ieeenat_fullname}
    \bibliography{main}
}

\clearpage
\setcounter{page}{1}
\maketitlesupplementary

\renewcommand{\thesection}{\Alph{section}}
\setcounter{section}{0}  

\renewcommand{\thefigure}{S.\arabic{figure}}
\setcounter{figure}{0} 

\renewcommand{\thetable}{S.\arabic{table}}
\setcounter{table}{0}

\section{Extracting Joint Coordinates: Model Setup}
\label{S.A}

In this study, we used the SwingNet architecture, which is a combination of Convolutional Neural Networks (CNN) and Recurrent Neural Networks (RNN) with a lightweight deep neural network architecture. The goal of this architecture is to detect and predict human postures during golf swing events. To detect eight key golf swing events, we employed a sequence mapping model fine-tuned by GolfDB and CaddieSet, which achieved an accuracy of 78.0\%. This represents an improvement of approximately 2\% over the baseline SwingNet model. Notably, accurately labeling the Address and Finish events was particularly challenging, as previously noted in \cite{mcnally2019golfdb}. This challenge arises due to the subjective nature of labeling and the inherent difficulty in precisely localizing these events temporally. After excluding the Address and Finish events, the accuracy for detecting the remaining six events increased to 94.1\%, compared to 91.8\% with the vanilla SwingNet model. Additionally, when testing on the MS COCO val2017 dataset \cite{lin2014microsoft}, the Faster R-CNN detector achieved a human average precision (AP) of 56.4. However, when paired with HRNet for pose estimation, the AP increased significantly to 74.9, which demonstrates the robustness of the models used for joint coordinate extraction, ensuring that the extracted joint data is reliable for further analysis.

\section{Detailed Feature Analysis for Swing-Related Metrics}
\label{S.B}

Based on domain knowledge, we utilized 15 swing-related metrics to extract features for each of the eight golf swing events. This allowed us to generate a total of 40 swing-related features. These features capture the joint angles and movements of various body parts during the swing, with each feature assigned to a specific phase of the swing. For example, the STANCE-RATIO is measured during the Address phase to capture the golfer's initial stance, while HIP-ROTATION is measured up to the Impact phase to track pelvic rotation.
These 40 features were analyzed in relation to BallSpeed and other key metrics, and the influence of each feature on the swing performance was evaluated. 

\subsection{Key Swing Metrics and Their Role}
\label{S.B.1}

The 15 key metrics are crucial for understanding the dynamics of a golf swing. For example, metrics such as SHOULDER-ANGLE, HIP-ROTATION, and WEIGHT-SHIFT are vital for generating power and controlling the swing's direction. Meanwhile, HEAD-LOC, SHOULDER-LOC, and STANCE-RATIO ensure that the swing remains consistent and accurate, facilitating better ball striking.

\subsection{Visualization of Features in the Swing Process}
\label{S.B.2}

We present a detailed analysis of how these features interact and affect ball trajectory. The heatmap in Figure \ref{fig:fig5} demonstrates the correlation between swing-related features and BallSpeed, showing that features such as STANCE-RATIO and SHOULDER-ANGLE are strongly correlated with ball speed and distance.

\section{Benchmark Evaluation: Model Setup}
\label{S.C}

For benchmarking, we employed various vision-based models including ResNet18, MobileNet\_V3, and ViT-B/16. These models were chosen for their widespread use in computer vision tasks and their expected strong performance when adapted to our swing videos. Specifically, we concatenated image patches from eight swing sequences into a single input, forming a combined \(320 \times 160\) input image, which was processed by the models for prediction. Each image represents one stage of the swing, capturing the progression over time.
The benchmark models were evaluated using various metrics such as Accuracy (Acc), Area Under the ROC Curve (AUC), and Mean Squared Error (MSE), as shown in the table \ref{tab:table_s3}.

\begin{table}[htb!]
    \centering
    \caption{Hyperparameter settings for the machine learning methods used in the experiments.}
    \resizebox{1.0\linewidth}{!}{%
    \begin{tabular}{cl}
        \hline
        \textbf{Methods}          & \textbf{Hyperparameters}                                                            \\ \hline
        \textbf{LR}               & penalty=`l2', C=1.0, solver=`lbfgs'                                               \\
        \textbf{SVM}              & C=1.0, kernel=`rbf', probability=True                                       \\
        \textbf{RF}               & n\_estimators=100, max\_features=`sqrt'                                               \\
        \textbf{XGBoost}          & use\_label\_encoder=False, eval\_metric=`logloss'                                       \\ \hline
    \end{tabular}%
    }
    \label{tab:table_s3}
\end{table}

\section{Down-the-Line (DTL) View Analysis}
\label{S.D}

\subsection{Data Collection from the DTL View}
\label{S.D.1}

\begin{figure*}[htb!]
    \centering
    \includegraphics[width=0.9\linewidth]{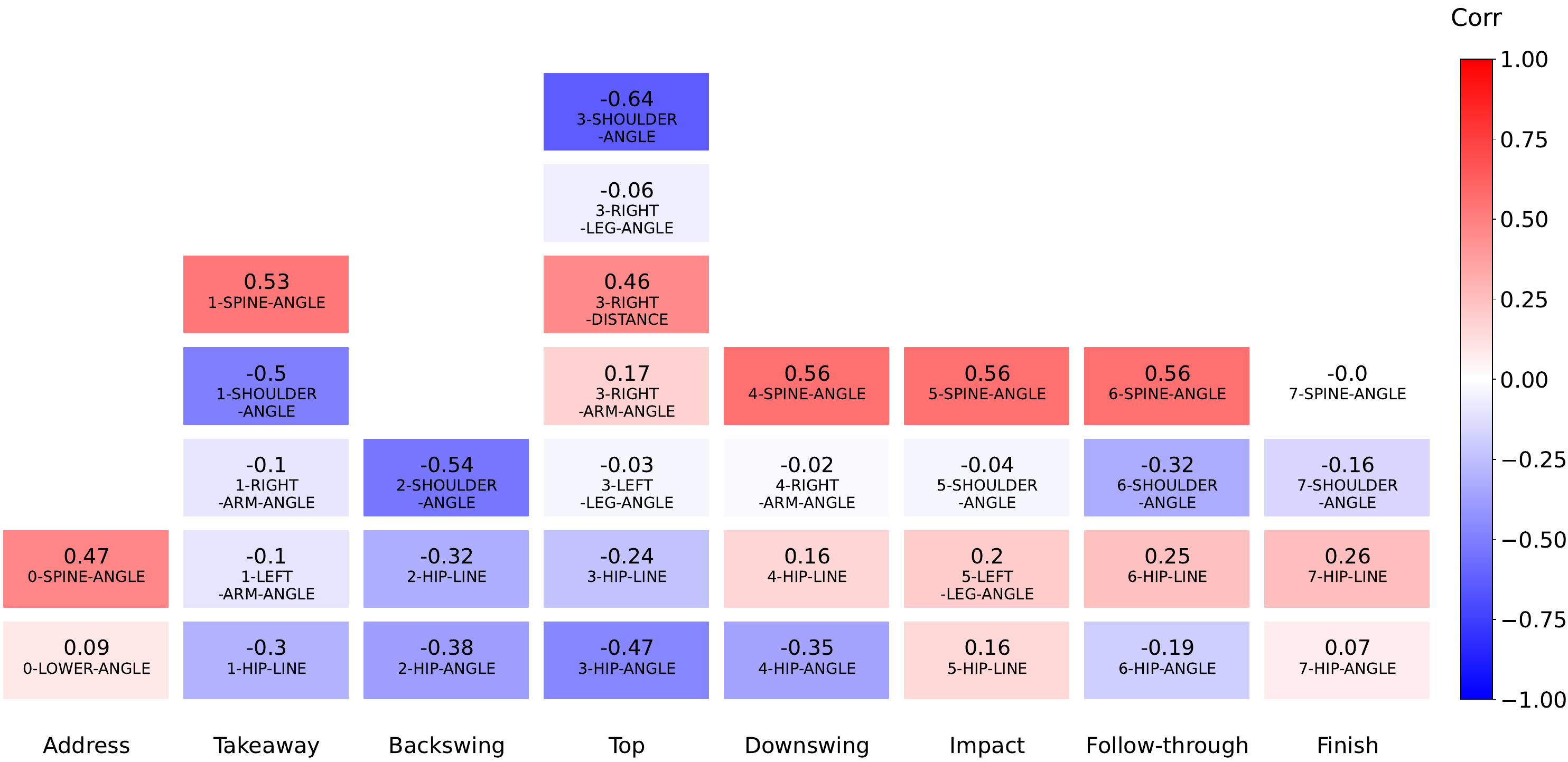}
    \caption{Correlation heatmap between swing-related features from \textsf{DTL} view and \textsf{BallSpeed}. For example, \texttt{1-SPINE-ANGLE} shows a significant positive correlation of 0.53. This angle, measured during the \textit{Takeaway}, is crucial for maintaining a stable swing plane and generating torque, contributing to increased ball speed. Additionally, hip-related features such as \texttt{3-HIP-ANGLE} and \texttt{4-HIP-LINE} show substantial correlations with ball speed, highlighting their importance in ensuring effective hip rotation and position during the swing.}
    \label{fig:fig_s1}
\end{figure*}

For a more comprehensive analysis, we also collected data from the Down-the-Line (DTL) view. This view allows us to capture key joint movements from a different perspective, providing additional insights into the swing mechanics. To extract swing-related features from the DTL view, we utilized nine metrics, which are table \ref{tab:table_s1}. The swing-related features extracted from the \textsf{DTL} view also demonstrate a close relationship with ball information, as illustrated in Figure \ref{fig:fig_s1}

\begin{table*}[htb!]
    \centering
    \caption{Description of the 9 metrics for \textsf{DTL} view: Each indicator measures specific joint information in the golf swing. These metrics help evaluate and improve the efficiency and effectiveness of the swing. They are based on insights from domain experts.}
    \resizebox{0.9\linewidth}{!}{
    \begin{tabular}{lll}
        \hline
        \textbf{Feature}                    & \textbf{Description}                                          & \textbf{Measurement} \\ \hline
        \texttt{SPINE-ANGLE}                & Spine angle relative to horizontal                            & degree   \\
        \texttt{LOWER-ANGLE}                & Angle formed by right pelvis, knee, and ankle                 & degree    \\
        \texttt{SHOULDER-ANGLE}             & Shoulder angle relative to horizontal                         & degree    \\
        \texttt{LEFT-ARM-ANGLE}             & Angle formed by left shoulder, elbow, and wrist               & degree    \\
        \texttt{RIGHT-ARM-ANGLE}            & Angle formed by right shoulder, elbow, and wrist              & degree    \\
        \texttt{HIP-LINE}                   & Movement of hip relative to Address                           & ratio   \\
        \texttt{HIP-ANGLE}                  & Rotation degree of pelvis relative to Address                 & degree   \\
        \texttt{RIGHT-DISTANCE}             & Gap between right elbow and the torso                         & ratio   \\
        \texttt{LEFT-LEG-ANGLE}             & Angle formed by left pelvis, knee, and ankle                  & degree   \\ \hline
    \end{tabular}%
    }
    \label{tab:table_s1}
\end{table*}

\subsection{DTL View Experimental Results}
\label{S.D.2}

CaddieSet for the DTL view comprises 833 samples, with 666 samples used for training and 167 for testing. Benchmarking these features showed significant correlations with ball speed, as detailed in the table \ref{tab:table_s2}.

\begin{table*}[htb!]
    \centering
    \caption{Benchmark comparison of model performance on the target variables \textsf{DirectionAngle}, \textsf{SpinAxis}, and \textsf{BallSpeed} on \textsf{DTL} view. The best experimental results are in bold.}
    \resizebox{0.7\textwidth}{!}{%
    \begin{tabular}{cccccc}
    \hline
    \multirow{2}{*}{\textbf{Method}} & \multicolumn{2}{c}{\textsf{DirectionAngle}} & \multicolumn{2}{c}{\textsf{SpinAxis}} & \textsf{BallSpeed}    \\ \cline{2-6} 
                                     & \textbf{Acc}     & \textbf{AUC}    & \textbf{Acc}  & \textbf{AUC} & \textbf{MSE} \\ \hline
    \textbf{ResNet18}                & \textbf{0.9162}        & 0.8190                & 0.7186              & 0.7200             & 74.74             \\
    \textbf{MobileNet\_V3}            & 0.8922                 & 0.7962                & \textbf{0.7246}     & 0.7267             & 37.61             \\
    \textbf{ViT-B/16}                     & 0.8802                 & 0.7648                & 0.6527              & 0.5712             & 48.27             \\ \hline
    \textbf{LR}                      & 0.9102                 & 0.8483                & 0.7126              & 0.7376             & 10.27             \\
    \textbf{SVM}                     & 0.8743                 & 0.8033                & 0.6527              & 0.7214             & 44.12             \\
    \textbf{RF}                      & 0.9042                 & 0.8366                & 0.7186              & 0.7176             & \textbf{7.35}     \\
    \textbf{XGBoost}                 & 0.9042                 & 0.8418                & 0.6946              & 0.7248             & 8.25             \\
    \textbf{NAM}                     & 0.8623                 & \textbf{0.9025}       & 0.6048              & \textbf{0.7434}    & 9.12             \\ \hline
    \end{tabular}%
    }
    \label{tab:table_s2}
\end{table*}

\end{document}